\documentclass{article}
\usepackage{spconf,amsmath,graphicx}
\usepackage{amsmath,graphicx}
\usepackage{color}

\usepackage{afterpage}
\usepackage{tabu}
\usepackage{tabularx}

\usepackage{makecell}
\let\OLDthebibliography\thebibliography
\renewcommand\thebibliography[1]{
  \OLDthebibliography{#1}
  \setlength{\parskip}{0pt}
  \setlength{\itemsep}{0pt plus 0.3ex}
}

\title{\LARGE{Improving 3D Occupancy Prediction through Class-balancing Loss and Multi-scale Representation}}
\name{Huizhou Chen$^{1,2,3}$, Jiangyi Wang$^{1,4}$, Yuxin Li$^{3,5}$, Na Zhao$^{4}$, Jun Cheng$^{1}$, Xulei Yang$^1$ \thanks{This research work is supported by the Agency for Science, Technology and Research (A*STAR) under its MTC Programmatic Funds (Grant No. M23L7b0021).}}
 \address{
 $^1$ Institute for Infocomm Research (I$^2$R), A*STAR,  Singapore \\
 $^2$ National University of Singapore (NUS), Singapore\\
  $^3$ Desay SV Automotive Singapore Pte. Ltd., Singapore \\
 $^4$ Singapore University of Technology and Design (SUTD), Singapore \\
 $^5$ Nanyang Technological University (NTU), Singapore
 }
\begin{document}
%
\maketitle
\begin{abstract}

3D environment recognition is essential for autonomous driving systems, as autonomous vehicles require comprehensive understanding of surrounding scenes. Recently, the predominant approach to define this real-life problem is through 3D occupancy prediction. It attempts to predict the occupancy states and semantic labels for all voxels in 3D space, which enhances the perception capability.
Bird’s-Eye-View(BEV)-based perception has achieved the SOTA performance for this task. Nonetheless, this architecture fails to represent various scales of BEV features. 
In this paper, inspired by the success of UNet in semantic segmentation tasks, we introduce a novel UNet-like Multi-scale Occupancy Head module to relieve this issue. Furthermore, we propose the class-balancing loss to compensate for rare classes in the dataset. The experimental results on nuScenes 3D occupancy challenge dataset show the superiority of our proposed approach over baseline and SOTA methods. 

\end{abstract}
%



\section{Introduction}
\label{sec:intro}
Accurately and comprehensively understanding of 3D environments is a fundamental part for autonomous driving systems \cite{mao20233d, TANG2023126587}. 
Traditional solutions define this task as 3D object detection \cite{7780605, huang2022bevdet, rukhovich2021imvoxelnet, yin2021centerbased}, 
which utilizes 3D bounding boxes to illustrate the information of surrounding objects.
However, this kind of definition has many drawbacks. 
Firstly, it struggles to accurately capture the intricate shapes of objects, leading to a loss of detailed information. Furthermore, the perception of background is absent, which can be useful for other downstream tasks like driving planning \cite{katrakazas2015real}.

As another definition of the perceiving task, 3D occupancy prediction strives to estimate the intricate occupancy state within a 3D space accurately. Here, occupancy refers to the status of individual voxels within a predefined area \cite{ConvOccNet}. 
It involves determining whether each voxel is occupied or unoccupied. If occupied, the goal is to classify the specific object or entity that occupies that particular voxel.
This task holds great importance within the domain of autonomous driving systems, as it significantly enhances the ability to perceive structures of various objects and regions. 
The input modality of this task is solely based on images from multiple cameras. NuScenes \cite{nuscenes} is a comprehensive and widely recognized dataset in the field of autonomous driving and computer vision. Within the nuScenes dataset, input images are sourced from six different cameras. For this specific dataset, the occupancy prediction is aimed to generate occupancy status results with dimension of $200\times200\times16$, which represents the realistic 3D space within the range of $[-40\text{m}, 40\text{m}] \times [-40\text{m}, 40\text{m}]\times [-1\text{m}, 5.4\text{m}]$.
When applied to autonomous driving scenarios, this definition serves the crucial purpose of scene comprehension, contributing significantly to safety enhancement. This is achieved by accurately forecasting the occupancy status of all voxels within the scene based on multi-view images.



Some existing solutions \cite{huang2023triperspective, wei2023surroundocc}
utilize a 3-stage paradigm to solve this problem. 
Firstly, it employs an image backbone for shallow feature extraction,and is followed by an image neck to fuse extracted features in different scales. Subsequently, the BEV feature encoder is designed to further transform the fused features into BEV space. This is particularly advantageous because BEV features are encoded in the top-down view of the 3D space, which naturally align with the data representation of the occupancy state, enabling them to encapsulate relevant features for voxels at various heights within their feature channels. Finally, it generates ultimate occupancy predictions from BEV features using an occupancy head.

In this study, we follow this paradigm and conduct experimental analysis.
We utilize BEV-based perception \cite{li2022bevformer} as the BEV feature encoder, and a simple Feed-Forward Networks(FFNs) occupancy head \cite{3DOccupancy2023} to generate baseline results. 
However, due to the fact that substantial imbalances exist among these categories within the dataset, this baseline method will significantly suffer from class-imbalanced issues. To this end, we introduce a class-balancing loss to tackle this issue. To further enhance feature representations, we develop a novel UNet-like Multi-scale Occupancy Head module, which is supervised via various scale ground truths.


\section{Related Work}
\label{sec:format}

\subsection{Voxel-based representation}


Attaining a comprehensive representation of a 3D scene constitutes a crucial step in understanding a lifelike environment. 
VoxelNet \cite{zhou2017voxelnet} partitions 3D space into equally spaced voxels and encoded the features within each voxel into a vector. 
In tasks involving occupancy prediction, it is crucial to represent the scene using voxel-based representations, with each voxel assigned a semantic label.
MonoScene \cite{DBLP:journals/corr/abs-2112-00726} stands out as the pioneering solution, which employs RGB images to reconstruct 3D scenes through voxel representations.
However, relying on information from a single viewpoint has inherent limitations. It will often lead to unsatisfactory prediction outcomes.
TPV-Former \cite{huang2023triperspective} leverages multi-view images as inputs, encoding features within a tri-perspective space. Subsequently, it generates semantic occupancy predictions via a dedicated prediction head. 
Although TPV-Former demonstrates significant improvement over other monocular solutions, its performance is hampered by the sparsity of Light-Detection-and-Ranging(LiDAR)-based ground truth data.

\subsection{BEV-based perception}
The substantial number of voxels poses a significant challenge to the computational efficiency for voxel-based approaches. To address this issue, BEV encoding methods leverage the relative scarcity of data in vertical dimension to integrate height details within each BEV grid. 
Furthermore, our occupancy prediction task can benefit from these existing, well-designed solutions for BEV feature encoding using a occupancy head. As height information is embedded within feature channels of the BEV gird, we are still able to capture a comprehensive description of the entire 3D space.
BEVFormer \cite{li2022bevformer} is a powerful transformer-based \cite{vaswani2017attention} BEV encoder. This encoder utilizes spatial-temporal attention mechanisms to achieve unified BEV feature representations. However, it does entail a higher computational load due to the in corporation of its attention layer.
BEVDet \cite{huang2021bevdet} employs Lift-Splat-Shoot (LSS) \cite{philion2020lift} module to perform the transformation from 2D to 3D space. This approach demands fewer computational resources owing to the omission of attention operations.
BEVDet4D \cite{huang2022bevdet4d} further enhances the performance by introducing temporal clues from BEV features of the previous frame.

\section{Methodology}
\label{sec:methodology}

\begin{figure}[t]
        \centering
        \includegraphics[width= 0.49\textwidth]{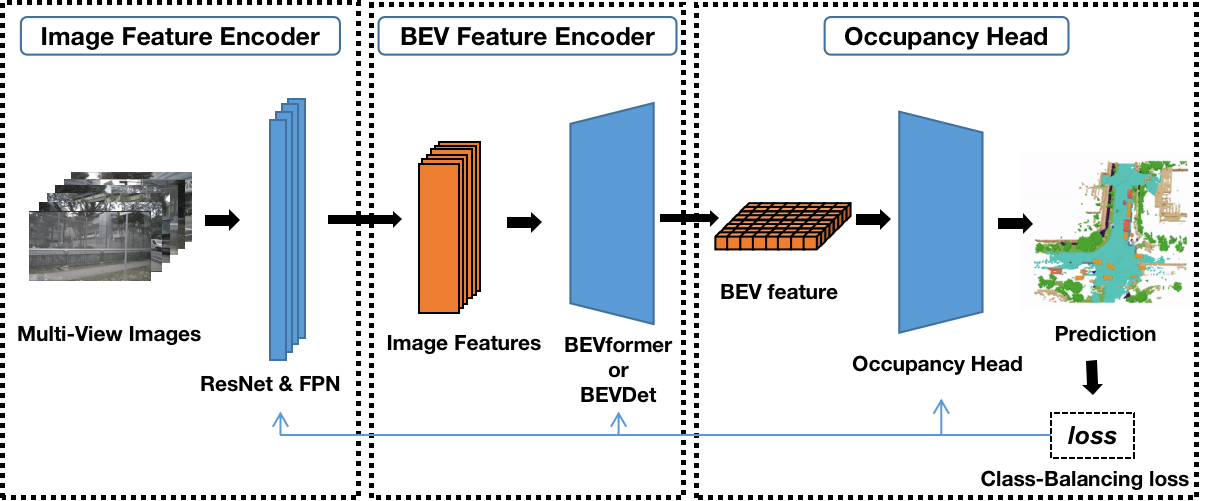}
        \caption{Overall Architecture of BEV-based Occupancy Prediction Model.}
        \label{fig: archi}
\end{figure}

In this section, we will describe the methodology employed in the 3D occupancy prediction experiments. The diagram of the overall architecture of a BEV-based occupancy prediction model is depicted in Fig. \ref{fig: archi}.
Firstly, the multi-view images are fed into an image feature encoder module for image feature extraction. This component utilizes a ResNet \cite{resnet} as its foundational backbone and complements it with an LSS-FPN \cite{huang2021bevdet} to serve as the image neck. 
Subsequently, BEV feature encoder module further processes these extracted image features into BEV features. Within our methodology, we leverage the capabilities of the BEVDet series \cite{huang2022bevdet4d}, known for their heightened efficacy and efficiency as compared to the BEVFormer \cite{li2022bevformer}.
Furthermore, we develop a novel occupancy head for occupancy prediction results. The naive occupancy head in \cite{3DOccupancy2023} entails straightforward feed-forward layers. Inspired by the achievements of the UNet \cite{cicek20163d} architecture in semantic segmentation tasks, we propose a novel UNet-like Multi-Scale Occupancy Head module to for better feature representations.
Lastly, to address the class-imbalanced issue in occupancy prediction, we introduce a novel class-balancing loss. This approach effectively encourages our networks to allocate more attention to the rare classes.
    


To conclude, our main contributions are:
\begin{itemize}
    \item \textbf{UNet-like Multi-scale Occupancy Head:} Different from naive FFNs occupancy head, we exploit the UNet architecture to enhance BEV feature representations. Furthermore, this allows us to supervise the whole training process with multi-scale ground truths. 
    \item \textbf{Class-balancing Loss:} We carefully design the novel loss function, which is able to compensate for minor classes within the dataset.
\end{itemize}

    \begin{figure*}[t]
        \centering
        \includegraphics[width=0.98\textwidth]{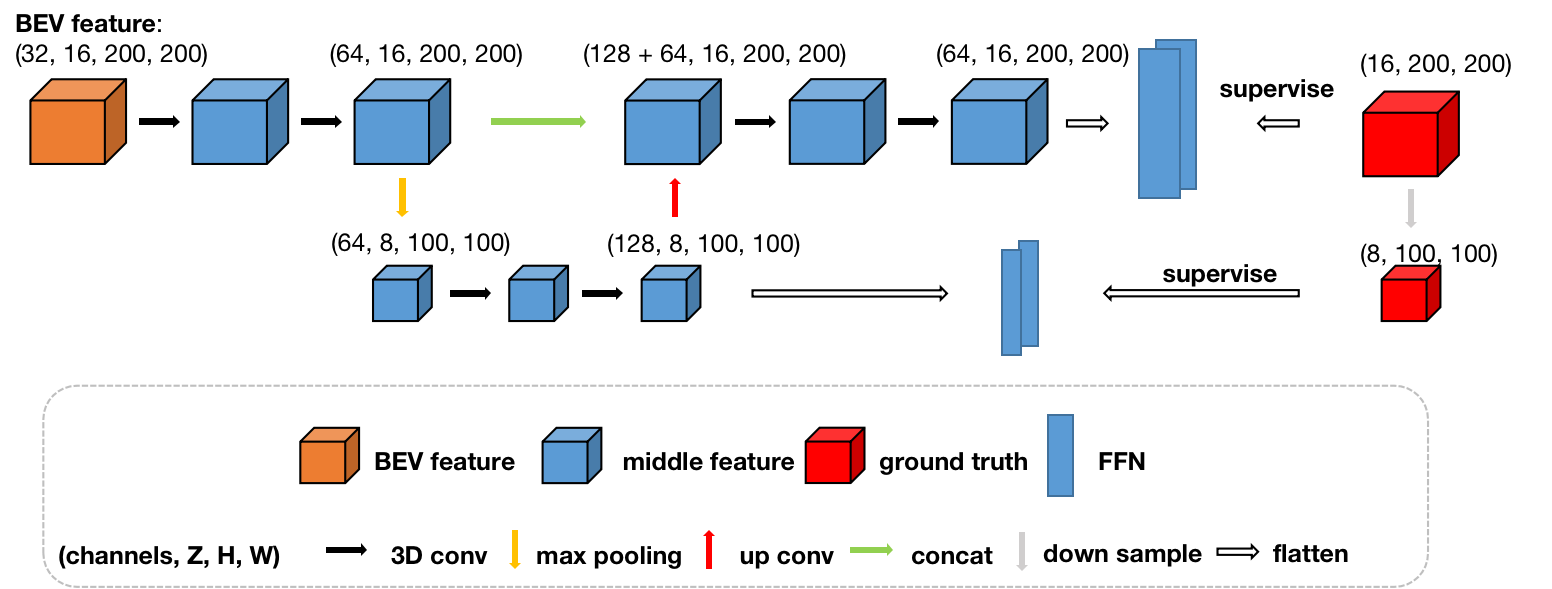}
        \caption{UNet-like Multi-Scale Occupancy Head. The architecture of proposed occupancy head within one layer. The arrows, distinguished by various colors, symbolize distinct operations, while the numbers enclosed in parentheses denote the dimension of the data, batch size is omitted here.}
        \label{fig: unet-l}
    \end{figure*}

     \subsection{UNet-like Multi-scale Occupancy Head}

    
    The occupancy head serves as a specialized module designed to utilize the BEV features for the prediction of occupancy states within a given scene. This module effectively leverages the BEV features to generate a comprehensive representation of spatial occupancy. Therefore, it allow us to provide crucial insights for tasks such as object detection and scene understanding.

    Assuming the BEV features are in the shape of [channels, Z, H, W]. The occupancy head used in the baseline model is a feed-forward network. It will directly flatten the BEV features into shape  [voxel\_nums, channels], followed by a linear layer to predict occupancy states.

    Considering the data format of BEV features is volumetric, we are inspired from 3D UNet \cite{cicek20163d}. Since the subsequent occupancy prediction task bears a solid resemblance to 3D semantic segmentation, the 3D UNet could naturally adapt to this task. To this end, we propose our novel UNet-like occupancy head and the architecture is shown in Fig. \ref{fig: unet-l}. Within this figure, the UNet architecture allows us to supervise the result with a smaller ground truth sampled from original one. This design enables our model to focus on various scales of ground truth.
    
    Our proposed UNet-like Multi-scale head has following advantages: 


    \begin{itemize}
         \item \textbf{Preservation of spatial context}: Since it operates on 3D volumes, the 3D UNet can capture and utilize spatial context information in all three dimensions. 

         \item \textbf{Better feature representations}: With 3D convolutions, the network can learn more comprehensive and informative features from the input data, leading to improved representations of the underlying structures. 

         \item \textbf{Consistency in 3D analysis}: Some tasks require consistent 3D analysis across the entire volume. The 3D UNet facilitates such tasks by maintaining the 3D context throughout the network, ensuring consistent and meaningful results.
    \end{itemize}


    Limited by the computational resources and the time, We only designed a shallow head for quick verification. The proposed occupancy head could be easily extended to 2 or 3 more layers by adding more convolution and de-convolution operations.


    \subsection{Class-balancing Loss}

\begin{table*}[t]
    \caption{Experimental results on nuScenes dataset show the superiority of our proposed method. The column labeled mIoU in the table represents the average IoU (Intersection over Union) of each class, while the columns named by class names represent the IoU of each class.\vspace{\baselineskip}}
    \label{table3}
    \centerline{
        \scalebox{0.75}{
        \begin{tabular}{c|c|c
        c@{\hspace{0.5em}}
        c@{\hspace{0.5em}}
        c@{\hspace{0.5em}}
        c@{\hspace{0.5em}}
        c@{\hspace{0.5em}}
        c@{\hspace{0.5em}}
        c@{\hspace{0.5em}}
        c@{\hspace{0.5em}}
        c@{\hspace{0.5em}}
        c@{\hspace{0.5em}}
        c@{\hspace{0.5em}}
        c@{\hspace{0.5em}}
        c@{\hspace{0.5em}}
        c@{\hspace{0.5em}}
        c@{\hspace{0.5em}}
        c@{\hspace{0.5em}}
        c@{\hspace{0.5em}}
        c@{\hspace{0.5em}}}
        \hline\hline
        Model & 
        Backbone & 
        mIoU & 
        \rotatebox{90}{others} & 
        \rotatebox{90}{barrier} & 
        \rotatebox{90}{bicycle} &
        \rotatebox{90}{bus} & 
        \rotatebox{90}{car} & 
        \rotatebox{90}{const. veh.} & 
        \rotatebox{90}{motorcycle} & 
        \rotatebox{90}{pedestrain} & 
        \rotatebox{90}{traffic cone} & 
        \rotatebox{90}{trailer} & 
        \rotatebox{90}{truck} & 
        \rotatebox{90}{drive. surf.} & 
        \rotatebox{90}{other flat} & 
        \rotatebox{90}{sidewalk} & 
        \rotatebox{90}{terrain} & 
        \rotatebox{90}{manmade} & 
        \rotatebox{90}{vegetation}\\\hline
        Baseline\cite{3DOccupancy2023} & 
        ResNet101 &
        23.68 &
        5.03 &
        38.79 &
        9.98 &
        34.41 &
        41.09 &
        13.24 &
        16.50 &
        18.15 &
        17.83 &
        18.66 &
        27.7 &
        48.95 &
        27.73 &
        29.08 &
        25.38 &
        15.41 &
        14.46 \\\hline
        BEVDet4D \cite{huang2022bevdet4d} & 
        ResNet50 &
        35.78 &
        7.84 &
        43.55 &
        15.19 &
        39.03 &
        48.92 &
        21.88 &
        19.42 &
        20.32 &
        21.46 &
        29.64 &
        35.04 &
        79.64 &
        37.26 &
        50.25 &
        52.86 &
        46.0 &
        39.94\\
        with loss & 
        ResNet50 &
        36.12 &
        10.76 &
        40.86 &
        24.64 &
        37.61 &
        47.09 &
        26.13 &
        25.51 &
        24.91 &
        25.58 &
        27.52 &
        34.2 &
        77.44 &
        37.74 &
        47.31 &
        51.18 &
        41.13 &
        34.39\\
        with loss\&head &
        ResNet50 &
        37.31 &
        11.2 &
        44.07 &
        24.31 &
        39.05 &
        48.68 &
        26.66 &
        25.48 &
        26.43 &
        28.9 &
        25.79 &
        36.91 &
        79.01 &
        38.79 &
        48.23 &
        52.01 &
        42.48 &
        36.26 \\
        \hline
        Improvement &
        \textbackslash &
        13.63 &
        6.17 &
        5.28 &
        14.33 &
        4.64 &
        7.59 &
        13.42 &
        8.98 &
        8.28 &
        11.07 &
        7.13 &
        9.21 &
        30.06 &
        11.06 &
        19.15 &
        26.63 &
        27.07 &
        21.8\\
        \hline
        \end{tabular}
        }
    }
\end{table*}

    The original loss function in BEV-based prediction model is a simple cross-entropy loss. The nuScenes dataset displays a considerable class imbalance, where standard classes like driveable surface and free occur approximately $10^{4}$ times more frequently than rare classes such as bicycle and motorcycle. However, cross-entropy loss will not account for class imbalances, leading the model to be biased toward predicting the majority class. To tackle this problem, we utilize weighted cross-entropy and dice loss to supervise the occupancy prediction. The weighted cross-entropy Loss assigns different weights to each class based on their prevalence in the dataset. The weights are typically inversely proportional to the class frequencies. In other words, the rarer the class, the higher the weight. While dice loss can effectively address class imbalance issues in image segmentation tasks due to the way it measures the overlap between predicted and true segmentation masks.

    To tackle computation challenges arising from a large number of voxel samples, we normalize the sample count across different classes. The final prediction loss is a combination of various components, including the weighted cross-entropy loss ($L_{wce}$), dice loss ($L_{dice}$), and depth loss for depth classification ($L_{depth}$) in LSS \cite{philion2020lift}. The overall loss function is described in the equation below:
    \begin{equation}
        Loss = w_{wce}L_{wce}+  w_{dice}L_{dice}+  w_{depth}L_{depth},
    \end{equation}
    where the loss weights $w_{wce}$, $w_{dice}$ and $w_{depth}$ are set to 1.0, 0.3 and 0.05 respectively based on experience.

\section{Experiments}

\vspace{-0.3in}
\begin{figure}[t]
        \centering
        \includegraphics[width=0.49\textwidth]{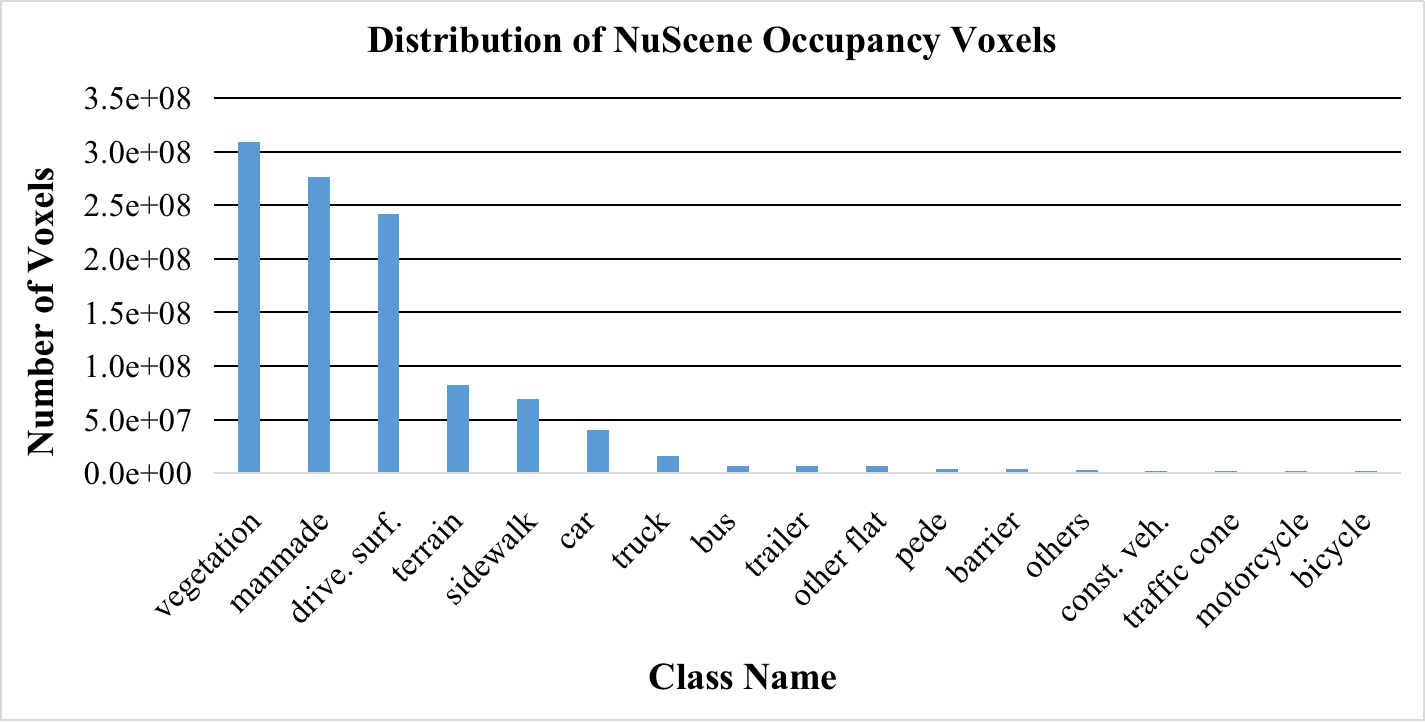}
        \vspace{-0.3in}
        \caption{Distribution of nuScenes Occupancy Voxels: The x-axis represents the class names, and the y-axis displays the cumulative voxel counts across all samples within the dataset.}
        \label{fig: data}
\end{figure}

    \subsection{Dataset}


    The official dataset employed in this study is the multi-view images from the nuScenes dataset \cite{nuscenes}. This dataset encompasses 1,000 scenes in reality, each equipped with 360-degree camera views and 32-beam LiDAR data, spanning a cumulative distance exceeding 1,000 kilometers. Within the multi-view image dataset, there are 28,130 samples allocated to the training split, 6,019 samples designated for the validation split, and 6,006 samples reserved for the test split. Each sample comprises six images, captured by six distinct cameras oriented as follows: front, rear, front-left, front-right, rear-left, and rear-right. Furthermore, the ground truth semantic labels encompass a total of 18 classes. Class 0 to class 16 correspond to different object categories that appear within the multi-view images. Class 17, denoted as free, is attributed to voxels occupied by nothing and will not be counted in the final mIoU.

    Due to variations in object sizes, differences in occurrence frequencies, and the substantial portion of empty space within the dataset, the nuScenes dataset exhibits a significant data imbalance issue in terms of occupancy labels. The distribution of voxels in each class can be found in Fig. \ref{fig: data}.

    \subsection{Implementation}
  
    For the training phase, we adhere to the original BEVDet4D design, where the multi-level features from the backbone were fused using LSS-FPN \cite{philion2020lift} and the temporal information is extracted from 2 previous frames. The multi-view images are cropped to dimensions of $704 \times 256$ pixels. During augmentation, vertical flipping and random scaling within the range of [0.94, 1.11] are applied, following the original design. We set batch size equal to 4 on 1 GPU and utilize AdamW \cite{loshchilov2019decoupled} optimizer with default parameters. The initial learning rate and weight decay rate are set to $1e^{-4}$ and $1e^{-2}$ respectively.

    \subsection{Experimental results}


    The baseline model we choose is BEVFormer \cite{3DOccupancy2023}. The image backbone ResNet101 and the heavy attention operations in the original BEVFormer bring a huge computation burden.  
    

In our experiments, we utilize BEVDet4D \cite{huang2022bevdet4d} as the BEV encoder. The experimental occupancy prediction results using BEVDet4D are presented in Table \ref{table3}.   
Initially, we directly employ BEVDet4D as the BEV feature encoder, utilize naive FFNs for the occupancy head, and apply cross-entropy loss. Remarkably, by simply adapting BEVDet4D with ResNet-50 as the backbone for this task, it exhibits a substantial performance improvement over the baseline, which employs ResNet-101 as the image backbone. 
Subsequently, we introduce our class-balancing loss function, including weighted cross-entropy, dice loss, and the original depth loss for the depth network in LSS \cite{philion2020lift}. Notably, these modifications yield improvements in certain small and infrequent classes, such as bicycle and motorcycle. 
Finally, we used both modified loss and the proposed UNet-like Multi-scale Occupancy Head, achieving the most favorable results among all conducted experiments. The last row in the table signifies the overall improvement observed from the baseline to our best result.
    

    



\section{Conclusion}

In this paper, our work has demonstrated substantial advancements for 3D occupancy prediction task.
We fix BEVDet4D as our BEV feature encoder, which is able to produce high-quality BEV feature representations. Moreover, we introduce a novel class-balancing loss to alleviate the class imbalance issue. Additionally, we propose a UNet-like Multi-Scale Occupancy Head, enhancing the quality of our feature representations and leading to further performance improvements. The experimental results on nuScenes dataset illustrate the superior efficacy of our proposed method.

\bibliographystyle{IEEEbib}
\bibliography{refs.bib}

\end{document}